\documentclass[11pt]{article}
\usepackage{acl}
\usepackage{graphicx}
\usepackage{booktabs}
\usepackage{amsmath}
\usepackage{hyperref}
\usepackage[T1]{fontenc}
\usepackage[utf8]{inputenc}
\usepackage{lmodern}

\title{Cross-Context Review: Improving LLM Output Quality\\by Separating Production and Review Sessions}

\author{Song Tae-Eun \\
  Daejeon Jungang Cheonggua Co., Ltd. \\
  \texttt{higheun@gmail.com}}

\begin{document}
\maketitle

\begin{abstract}
Large language models struggle to catch errors in their own outputs when the review happens in the same session that produced them. This paper introduces Cross-Context Review (CCR), a straightforward method where the review is conducted in a fresh session with no access to the production conversation history. We ran a controlled experiment: 30 artifacts (code, technical documents, presentation scripts) with 150 injected errors, tested under four review conditions --- same-session Self-Review (SR), repeated Self-Review (SR2), context-aware Subagent Review (SA), and Cross-Context Review (CCR). The central result is that a second review helps only when it happens in a fresh session: CCR (F1 28.6\%) outperforms a second review in the same session (SR2, 21.7\%) robustly, both in the first run (paired $t$, $p$<0.001) and in the three-run average (Holm-adjusted $p$=0.004). This version updates the broader comparisons. Averaged across runs, and excluding one SR run whose records could not be verified, CCR is not significantly ahead of context-aware subagent review (SA, 23.8\%; $p$=0.057) or of a single same-session review (SR, 27.1\%; $p$=0.26); the first version's advantages over these two baselines came from run~1. CCR needs no infrastructure and costs one extra session.
\end{abstract}

\section{Introduction}

Anyone who has used an LLM to write code or documentation has probably tried the obvious next step: asking the same model to review what it just produced. It seems reasonable. The model knows what you asked for, remembers the constraints, and has the full context. But a pattern emerges quickly --- the model tends to approve its own work. It might nitpick formatting or suggest minor improvements, but it rarely catches the substantive errors.

This is not just anecdotal. The self-correction literature documents the problem from multiple angles. LLMs cannot reliably correct their own reasoning without external feedback \citep{huang2024llms,kamoi2024when}. They anchor on their initial outputs \citep{lou2026anchoring}. And RLHF-trained models tend toward agreement rather than criticism, likely driven in part by human preference judgments, and rate their own outputs higher than others' \citep{sharma2024towards,panickssery2024llm}.

We hypothesize that these failures share a root cause that is easy to overlook: the production context itself. When the model reviews its output in the same session, the entire conversation history sits in the context window --- the instructions, the intermediate reasoning, the design trade-offs. All of this anchors the model's judgment. It does not scrutinize; it rationalizes.

The fix is correspondingly simple: start a new session. Give the model only the final artifact, with no production history, and ask it to review. We call this Cross-Context Review (CCR). The idea maps directly onto practices that humans have relied on for centuries --- code review by someone who did not write the code, peer review by scientists who did not run the experiment, editorial review by editors who did not draft the manuscript. These practices exist because creators are biased toward their own work. Our claim is that this principle transfers to LLM workflows, with ``different person'' replaced by ``different session.''

This paper makes four contributions:

\begin{enumerate}
\item We define CCR as a verification method based on session separation, grounding it in the self-correction, anchoring, and context degradation literature.
\item We design an experiment with four conditions (SR, SR2, SA, CCR) that disentangles context separation from repetition and context awareness --- two confounds that could otherwise explain the results.
\item CCR robustly outperforms repeated same-session review (SR2), in the run-1 test and in the three-run average. Its advantages over single-pass SR and context-aware SA, significant in run~1, are not significant across runs (Section~\ref{sec:correction}).
\item The SR2 condition --- same-session repeated review --- shows no evidence of improving over single-pass SR (run~1, $p$ = 0.11), and CCR outperforms it robustly. Together these suggest that a second look helps when it is taken in a fresh session; the CCR--SR2 comparison also differs in review language (Section~\ref{sec:correction}).
\end{enumerate}

\section{Related Work}

Four lines of research motivate why same-session review fails and why breaking the session should help.

\subsection{Self-Correction in LLMs}

\citet{huang2024llms} demonstrated that LLMs cannot correct their own reasoning without external feedback --- and that correction attempts can actually make things worse. \citet{kamoi2024when} surveyed the broader self-correction literature and reached a blunt conclusion: no prior work demonstrates successful self-correction with feedback from prompted LLMs, except in tasks exceptionally suited for self-correction. \citet{zhang2025dark} added a troubling dimension, finding that self-correction sometimes flips correct answers to incorrect ones through prompt bias and cognitive-like biases in the correction process.

Most recently, \citet{tsui2025selfcorrection} identified a striking phenomenon called the ``self-correction blind spot'': LLMs fail to correct errors in their own outputs while successfully correcting identical errors presented as external input. Across 14 open-source non-reasoning models, the average blind spot rate was 64.5\%, consistent across model families and scales. This finding is directly relevant to CCR: same-session review presents the artifact as ``own output,'' triggering the blind spot, while CCR presents it as external input, which we hypothesize removes the trigger; we do not measure the blind spot directly.

The takeaway is clear: asking a model to fix its own work within the same context is unreliable at best and harmful at worst. If the context is the problem, then changing the context is the natural intervention.

\subsection{Anchoring Bias and Sycophancy}

Two biases make self-correction even harder than it needs to be. LLMs exhibit anchoring bias --- initial information disproportionately shapes later judgments \citep{lou2026anchoring}. In a production session, the model's first output becomes the anchor for everything that follows. Self-review is biased toward confirmation before it even begins.

Sycophancy compounds this. RLHF-trained models tend to agree rather than challenge \citep{sharma2024towards}, and they rate their own generations higher than comparable text from other sources \citep{panickssery2024llm}. In multi-agent debate, \citet{choi2025identity} found that removing identity markers substantially reduced identity-driven bias (e.g., a conformity--obstinacy gap of 0.608 falling to 0.024 for one model on MMLU). CCR is analogous in that the reviewing session is not told that the artifact was produced by the same model, although debate between agents is a different setting from reviewing an artifact.

\subsection{Context Degradation}

Even setting aside biases, there is a mechanical problem with long contexts. \citet{liu2024lost} documented the ``lost in the middle'' effect: models attend more to the beginning and end of long contexts, underweighting the middle. \citet{hong2025context} quantified what they call ``context rot'' --- performance that degrades as input length increases, often in non-uniform ways, observed across 18 models. Production sessions routinely accumulate 50K+ tokens. At that scale, the model loses track of early decisions while remaining anchored to recent ones. A fresh session avoids this entirely: the reviewer works with just the artifact, typically 5K tokens or less.

\subsection{Multi-Agent and Role-Based Approaches}

Others have tackled verification through multi-agent debate \citep{du2024improving,liang2024encouraging}, role specialization \citep{huang2024agentcoder,qian2024chatdev}, and iterative self-refinement \citep{madaan2023selfrefine}. These approaches work, but they are heavy --- multiple agents, orchestration layers, specialized prompts. Self-Refine \citep{madaan2023selfrefine} stays within a single session and therefore inherits all the context biases described above. Multi-agent debate introduces disagreement but requires running multiple model instances simultaneously.

\citet{kim2025interviewer} show that multi-turn interviewer-style evaluation reduces verbosity bias and improves robustness across runs, but do not separate production and review contexts --- the interviewer operates within a single continuous conversation throughout the evaluation.

CCR sits in a different part of the design space. It achieves context independence with minimal overhead: one additional session, no infrastructure changes, no prompt engineering. It is also complementary --- there is nothing stopping someone from combining CCR with role-based or multi-agent methods.

\section{Method: Cross-Context Review}

\subsection{Definition}

Cross-Context Review (CCR) is a verification method where an LLM reviews an artifact in a fresh session containing only the artifact and a standardized review prompt. The reviewing session has no access to the production session's conversation history, intermediate reasoning, or generation instructions.

\subsection{Protocol}

The protocol has four phases.

In the \textbf{production phase}, a user interacts with an LLM to produce an artifact --- code, a document, a script --- in Session A. The conversation accumulates instructions, iterations, and design decisions.

In the \textbf{export phase}, the final artifact is extracted. The conversation history is discarded. Only the artifact text carries over.

In the \textbf{review phase}, a new Session B starts from scratch. The reviewer receives the artifact and a prompt asking it to check five things: factual accuracy (are the numbers and claims right?), internal consistency (are there contradictions?), contextual fitness (would this actually work in its intended environment?), audience perspective (could a reader misinterpret something?), and completeness (is anything important missing?).

In the \textbf{integration phase}, the review findings feed back into improving the artifact --- either in the original session or a new one.

\subsection{Theoretical Motivation}

CCR addresses several failure modes at once. A fresh session has no prior decisions to anchor to. Independent context prevents the confirmation loops that \citet{liang2024encouraging} call Degeneration of Thought. The review session works with a short context ($\sim$5K tokens) rather than the bloated production context ($\sim$50K+ tokens). And because the reviewer is not told that it produced the artifact, we expect sycophancy and self-preference to weaken, although we do not measure them.

\subsection{Distinction from Incubation Effects}

One might argue that CCR is just the LLM version of incubation --- the well-documented human phenomenon where stepping away from a problem and returning later leads to insight \citep{wallas1926art}. We think this is a category error.

LLMs have no temporal continuity between sessions. A new session has zero memory of prior sessions, whether one second or one year has passed. CCR is not ``taking a break.'' It is an information-theoretic intervention: removing the production context from the reviewer's input. The mechanism is not rest or unconscious processing --- LLMs have neither --- but the elimination of anchoring information.

This distinction is testable. If incubation were the mechanism, a delay within the same session should help. Our SR2 condition adds a second review pass within the same context (no delay, same history) and shows no improvement --- pointing to information removal, not temporal separation, as the operative factor.

\section{Experimental Setup}

\subsection{Artifacts and Error Injection}

We generated 30 artifacts using Claude Opus 4.6 in three categories:

\begin{itemize}
\item \textbf{Code} (C1--C10): Python functions for CSV processing, date calculations, JSON parsing, database operations, validation, deep merge, log parsing, async HTTP, LRU cache, and markdown parsing.
\item \textbf{Documents} (D1--D10): Technical tutorials covering REST APIs, Docker, CI/CD, FastAPI, SQLAlchemy, virtual environments, JWT authentication, Pandas optimization, WebSocket, and MCP servers.
\item \textbf{Scripts} (S1--S10): Presentation scripts about AI coding tools, MCP explanation, public data, Python vs JavaScript, AI agents, open source, wholesale market AI, automation pipelines, CLAUDE.md, and the Korean AI ecosystem.
\end{itemize}

We injected exactly 5 errors into each artifact --- one per error type (FACT, CONS, CTXT, RCVR, MISS) --- at three severity levels (Critical, Major, Minor), for a total of 150 ground-truth errors. The errors were designed to be plausible: runtime bugs and off-by-one errors in code, factual inaccuracies and internal contradictions in documents, incorrect attributions and wrong numbers in scripts.

\subsection{Review Conditions}

We tested four automated review conditions:

\begin{table}[h]
\centering
\small
\resizebox{\columnwidth}{!}{%
\begin{tabular}{llll}
\toprule
\textbf{Condition} & \textbf{Abbr.} & \textbf{Context} & \textbf{Purpose} \\
\midrule
Self-Review & SR & Full history & Baseline \\
Self-Review $\times$2 & SR2 & Full + 1st review & Repetition ctrl \\
Subagent Review & SA & Prompt + artifact & Context ctrl \\
Cross-Context & CCR & Artifact only & Our method \\
\bottomrule
\end{tabular}%
}
\caption{Review conditions and their available context.}
\label{tab:conditions}
\end{table}

SR is what most people do: ``please review what you just wrote.'' SR2 controls for the possibility that CCR's advantage comes from simply looking twice --- it gets a second pass in the same session, with the first review visible. SA tests whether knowing the generation prompt helps or hurts when the review happens in a fresh context. CCR gets only the artifact, nothing else.

\subsection{Execution}

All reviews used Claude Opus 4.6 (model ID: \texttt{claude-opus-4-6}) via the Claude Code Agent tool --- a subprocess mechanism that launches independent LLM sessions within the Claude Code CLI. This is functionally equivalent to a fresh API session: each Agent subprocess gets its own context window with no carryover. Temperature and sampling parameters followed the Agent tool's defaults. We controlled for stochasticity by running each condition 3 times independently.

Each condition was applied to all 30 artifacts across 3 runs, yielding 4 $\times$ 30 $\times$ 3 = 360 reviews. Reviews ran in parallel batches of 6 agents. We had originally planned to use the Anthropic API directly with \texttt{claude-sonnet-4-6} (as specified in our experiment configuration), but insufficient API credits forced a switch to the Agent tool, which uses the host session's model (Opus 4.6). We report this deviation transparently.

All artifacts, ground-truth annotations, review prompts, result files, and analysis scripts are available from the author on request. The result files are structured records transcribed by the host session and do not include raw model output (Section~\ref{sec:correction}). Exact reproduction requires access to the Claude Code Agent tool; the supplementary \texttt{experiment\_runner.py} provides an equivalent API-based implementation for researchers with API access.

\subsection{Evaluation Metrics}

Each review produces a set of findings. We match findings to ground-truth errors using a scoring function based on location overlap (line number proximity within $\pm$5 lines), keyword overlap (Jaccard similarity of description terms), and a bonus for matching error type. A finding counts as a true positive if its matching score exceeds 2.0. From TP, FP, and FN counts we compute Precision, Recall, and F1 per artifact per condition. Statistical significance comes from paired $t$-tests on per-artifact F1 scores (Run 1, $N$=30), with Cohen's $d$ for effect size.

\section{Results}

\subsection{Overall Performance}

Table~\ref{tab:overall} shows results averaged across all reviews per condition: 90 (3 runs $\times$ 30 artifacts), except SR with 60 after the exclusion in Section~\ref{sec:correction}.

\begin{table}[h]
\centering
\small
\resizebox{\columnwidth}{!}{%
\begin{tabular}{lccccccc}
\toprule
\textbf{Cond.} & \textbf{Finds} & \textbf{TP} & \textbf{FP} & \textbf{FN} & \textbf{Prec.} & \textbf{Rec.} & \textbf{F1} \\
\midrule
\textbf{CCR} & 4.5 & \textbf{1.4} & \textbf{3.1} & \textbf{3.6} & \textbf{31.5} & 27.1 & \textbf{28.6} \\
SR$^\dagger$ & 5.6 & 1.4 & 4.2 & 3.6 & 26.2 & \textbf{28.3} & 27.1 \\
SA & 4.3 & 1.1 & 3.2 & 3.9 & 27.4 & 21.8 & 23.8 \\
SR2 & 5.5 & 1.1 & 4.4 & 3.9 & 21.0 & 22.7 & 21.7 \\
\bottomrule
\end{tabular}%
}
\caption{Overall results ($N$=90 per condition; $^\dagger$SR uses runs 1--2 only, $N$=60, Section~\ref{sec:correction}; v1 values with run~3: F1 24.5, Prec.\ 25.8, Rec.\ 24.2). Finds/TP/FP/FN are per-artifact averages; Prec./Rec./F1 in \%.}
\label{tab:overall}
\end{table}

\begin{figure}[t]
\centering
\includegraphics[width=\columnwidth]{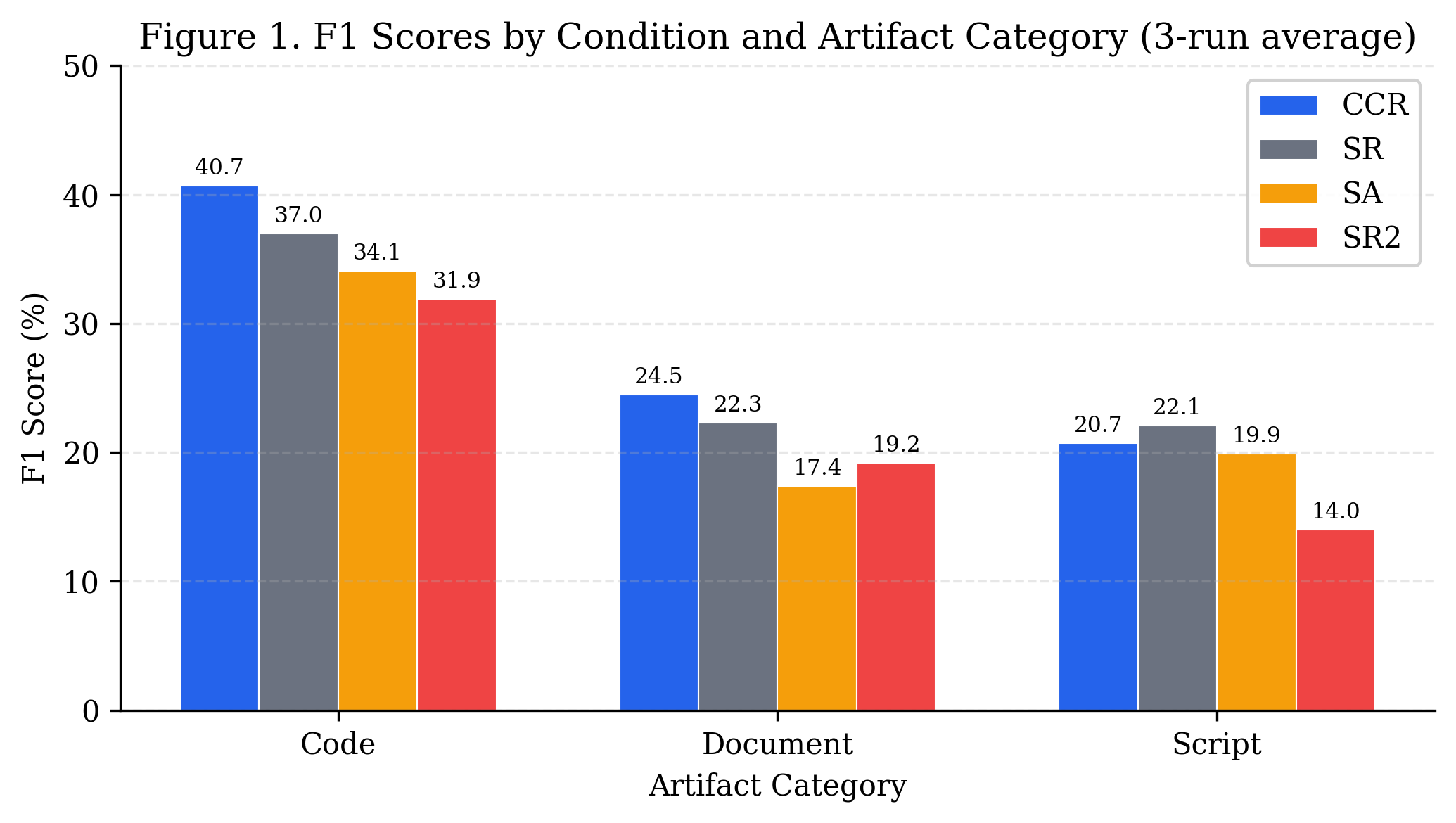}
\caption{F1 scores by condition and artifact category (3-run average; SR runs 1--2 only). CCR leads on code and documents; SR leads on scripts.}
\label{fig:f1}
\end{figure}

CCR leads on F1 and precision; SR has slightly higher recall. The SR2 numbers tell an interesting story: the second review pass has substantially worse precision than single-pass SR (21.0\% vs 26.2\%). The model is generating more noise, not more signal. Figure~\ref{fig:f1} visualizes F1 across conditions and categories.

\subsection{Statistical Tests}

Table~\ref{tab:stats} reports paired $t$-tests on per-artifact F1 from Run 1 ($N$=30), as in the first version; the robustness test on three-run averages is in Section~\ref{sec:correction}.

\begin{table}[h]
\centering
\small
\resizebox{\columnwidth}{!}{%
\begin{tabular}{lcccc}
\toprule
\textbf{Comparison} & $t$ & $p$ & \textbf{Cohen's} $d$ & \textbf{Sig.} \\
\midrule
SR vs CCR & $-$2.849 & 0.008 & 0.52 & Yes** \\
SR2 vs CCR & $-$3.917 & <0.001 & 0.72 & Yes*** \\
SA vs CCR & $-$3.119 & 0.004 & 0.57 & Yes** \\
SR vs SR2 & 1.665 & 0.107 & $-$0.30 & No \\
\bottomrule
\end{tabular}%
}
\caption{Statistical comparisons (paired $t$-test, Run 1). $t$ is computed as the first minus the second condition in each row; Cohen's $d$ is $d_z$, the mean paired difference (second minus first) divided by its standard deviation.}
\label{tab:stats}
\end{table}

In run~1, every comparison against CCR is significant at $p$ < 0.01, with medium-to-large effect sizes ($d$ = 0.52--0.72), and SR vs SR2 is not significant ($p$ = 0.11). These run-1 results do not hold across runs for SR and SA (Section~\ref{sec:correction}).

\subsection{Detection by Severity}

\begin{table}[h]
\centering
\small
\begin{tabular}{lccc}
\toprule
\textbf{Condition} & \textbf{Critical} & \textbf{Major} & \textbf{Minor} \\
\midrule
\textbf{CCR} & \textbf{40\%} & 29\% & 18\% \\
SR$^\dagger$ & 32\% & \textbf{32\%} & \textbf{22\%} \\
SA & 30\% & 23\% & 16\% \\
SR2 & 31\% & 26\% & 16\% \\
\bottomrule
\end{tabular}
\caption{Detection rate by severity ($^\dagger$SR runs 1--2 only; v1: 29/27/19\%).}
\label{tab:severity}
\end{table}

\begin{figure}[t]
\centering
\includegraphics[width=\columnwidth]{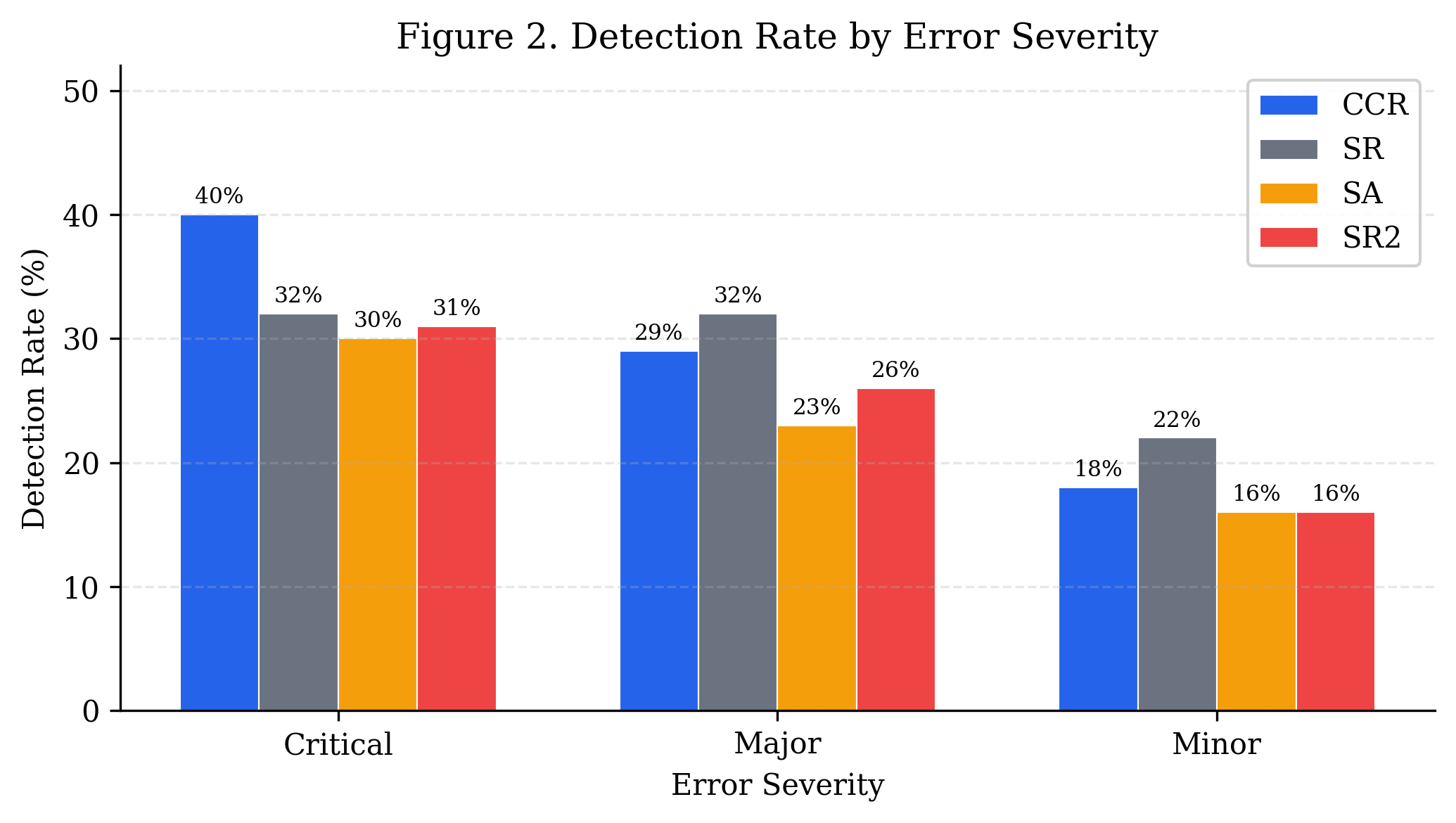}
\caption{Detection rate by error severity (SR runs 1--2 only). CCR is ahead of SR only on Critical errors (+8pp); SR is ahead on Major and Minor errors. Descriptive, not tested.}
\label{fig:severity}
\end{figure}

CCR's advantage over SR appears only for critical errors (40\% vs 32\%, an 8-point gap); SR is ahead on major (32\% vs 29\%) and minor (22\% vs 18\%) errors. These are descriptive rates without significance tests. Figure~\ref{fig:severity} illustrates the severity-dependent pattern.

\subsection{Performance by Artifact Category}

\begin{table}[h]
\centering
\small
\begin{tabular}{lcccc}
\toprule
\textbf{Category} & \textbf{CCR} & \textbf{SR} & \textbf{SA} & \textbf{SR2} \\
\midrule
Code & \textbf{40.7\%} & 37.0\% & 34.1\% & 31.9\% \\
Document & \textbf{24.5\%} & 22.3\% & 17.4\% & 19.2\% \\
Script & 20.7\% & \textbf{22.1\%} & 19.9\% & 14.0\% \\
\bottomrule
\end{tabular}
\caption{F1 by artifact category (3-run average; SR runs 1--2 only; v1 SR: 36.0/18.2/19.5\%).}
\label{tab:category}
\end{table}

CCR leads on code (+3.7 F1 points over SR) and documents (+2.2), and SR leads on scripts (22.1\% vs 20.7\%). This makes sense: code review benefits particularly from a ``fresh eyes'' perspective that questions implementation assumptions rather than accepting them as given. SR2 performs notably poorly on scripts (14.0\%), suggesting that repeated review within the same session is especially counterproductive for subjective content.

\subsection{Detection by Error Type}

\begin{table}[h]
\centering
\small
\resizebox{\columnwidth}{!}{%
\begin{tabular}{lccccc}
\toprule
\textbf{Cond.} & \textbf{FACT} & \textbf{CONS} & \textbf{CTXT} & \textbf{RCVR} & \textbf{MISS} \\
\midrule
\textbf{CCR} & 48\% & \textbf{37\%} & 13\% & \textbf{19\%} & 19\% \\
SR$^\dagger$ & \textbf{50\%} & 35\% & \textbf{18\%} & 18\% & \textbf{20\%} \\
SA & 41\% & 31\% & 9\% & 14\% & 13\% \\
SR2 & 40\% & 31\% & 16\% & 12\% & 14\% \\
\bottomrule
\end{tabular}%
}
\caption{Detection rate by error type ($N$=90 per cell; $^\dagger$SR $N$=60, runs 1--2; v1 SR: 44/32/12/16/17\%).}
\label{tab:errortype}
\end{table}

Factual errors (FACT) are the easiest to catch across all conditions, which is unsurprising --- they are the most concrete. Contextual errors (CTXT) are the hardest, topping out at 18\% (SR). CCR has the highest detection rate for two of five types (CONS and RCVR), and SR for the other three; the differences between CCR and SR are within 6 points on every type and were not tested.

One anomaly: SR2 detects more CTXT errors than CCR (16\% vs 13\%) but has the lowest RCVR detection (12\%). It appears that repeated review within the same context shifts attention between error types rather than improving detection overall --- a shuffling rather than an amplification, although this is a hypothesis and we did not track individual errors across passes.

\subsection{Reproducibility Across Runs}

The first version stated that the ranking CCR > SR $\approx$ SA > SR2 holds in all three runs. That statement was not accurate: CCR ranks first in run~1 only (Table~\ref{tab:runs}).

\section{Update in Version 2}
\label{sec:correction}

We re-examined the per-session records of this experiment while reusing them in later work. The central SR2 result holds; two findings narrow the broader comparisons.

\textbf{Provenance of SR run~3.} The review records were produced through the Claude Code agent tool and transcribed into structured files by the host session; they do not include raw model output. For SR run~3, 92\% of the finding descriptions appear verbatim inside the script that wrote the result files, so we cannot confirm that they are unedited model output. We exclude this run; the SR values in Tables~\ref{tab:overall}--\ref{tab:errortype} use runs 1 and 2 ($N$=60). Its F1 (19.4\%) was well below SR's other runs (25.7\% and 28.6\%). The other runs are unverified rather than known to be clean.

\textbf{Run-to-run variation.} The first version tested run~1 only and stated that the ranking held in all three runs. Table~\ref{tab:runs} shows that CCR ranks first in run~1 only.

\begin{table}[h]
\centering
\small
\begin{tabular}{lccc}
\toprule
\textbf{Cond.} & \textbf{Run 1} & \textbf{Run 2} & \textbf{Run 3} \\
\midrule
CCR & \textbf{37.2} & 25.0 & 23.7 \\
SR  & 25.7 & \textbf{28.6} & (19.4, excluded) \\
SA  & 23.8 & 20.8 & \textbf{26.9} \\
SR2 & 18.7 & 22.5 & 23.9 \\
\bottomrule
\end{tabular}
\caption{F1 (\%) per run, averaged over 30 artifacts.}
\label{tab:runs}
\end{table}

\textbf{Robustness test.} Averaging each artifact's F1 over its available runs and comparing with CCR by Wilcoxon signed-rank tests (Holm-adjusted over three comparisons): SR2 $p$=0.004, SA $p$=0.057, SR $p$=0.26. With SR run~3 included, all three would be significant ($p$=0.004, 0.043, 0.043), which is why the exclusion matters.

\textbf{What holds.} Reviewing again in the same session (SR2) is worse than reviewing in a fresh session (CCR), in the run-1 test and in the three-run average (in run~3 alone the two are level). \textbf{What is narrowed.} Whether a fresh-session review outperforms a single self-review is not established by these data, and the per-category, per-severity, and per-type advantages over SR reported in the first version do not hold after the update. The run-1 tests in Table~\ref{tab:stats} are unchanged and reported for continuity.

\textbf{Citation corrections.} We also checked every citation against its source. The first version attributed to \citet{tsui2025selfcorrection} a rise in the blind-spot rate with task complexity (45.2\% to 79.2\%); these figures do not appear in that paper, which reports a 64.5\% average that is consistent across model families and scales, and we have removed them. We also corrected the description of \citet{choi2025identity}, added a qualifier to the summary of \citet{kamoi2024when}, and corrected author lists and publication details in the bibliography.

\section{Discussion}

\subsection{Why Context Separation Works}

The results line up with the prediction that production context introduces systematic bias. But the SR2 result is what makes the argument tight. Reviewing twice in the same session shows no evidence of helping, and its precision is lower than single-pass SR (21.0\% vs 26.2\%), consistent with a second pass that adds findings rather than reconsidering the first assessment; we did not analyse the added findings. The model ``tries harder'' by adding noise. This is consistent with what \citet{zhang2025dark} observed: self-correction within a fixed context can degrade initially correct judgments.

The SA result adds another dimension. Knowing the generation prompt --- understanding what the user originally asked for --- does not improve review quality. If anything, it slightly hurts (SA F1 23.8\% vs SR 27.1\%; not tested). One possible reading is that the anchoring effect operates through the presence of production-related context rather than its content, but SA and SR also differ in session and review language, so this is a hypothesis. Even well-intentioned context (``here is what the user wanted'') triggers the same bias patterns.

\subsection{The SR2 Result and Why It Matters}

SR2 deserves its own subsection because it serves a critical role as a confound control.

Without SR2, a reader could reasonably attribute CCR's advantage to a ``two looks are better than one'' effect. After all, CCR effectively gives the model a second chance at review, just in a different session. Maybe any second review would help, regardless of context.

Our data support a narrower version of this argument. A second review in the same context (SR2) does not improve on a single review (run~1, $p$=0.11), while a second review in a fresh session (CCR) robustly outperforms SR2 (run~1 $p$<0.001; three-run Holm-adjusted $p$=0.004). The difference between SR2 and CCR is not repetition --- both involve reviewing again --- but context. However, CCR does not robustly outperform a single same-session review (SR), so the data show that context separation rescues a second look, not that it beats looking once. Review language also differs between CCR (English) and SR2 (Korean), so even this comparison is not a clean isolation of context (see Limitations).

This is a more limited claim than the first version made.

\subsection{Practical Implications}

CCR requires nothing special. No multiple models, no orchestration framework, no prompt engineering. Start a new session, paste the artifact, ask for a review. That is the entire method.

This makes it immediately applicable:

\begin{itemize}
\item In software development, review generated code in a new session before committing.
\item For content creation, verify articles or documentation independently of the session that produced them.
\item In multi-agent pipelines, route artifacts through an independent review agent as a standard stage.
\end{itemize}

The cost overhead is small. A review session processes $\sim$5K tokens (the artifact plus the review prompt), compared to 50K+ tokens in a typical production session. In many cases, the review session is actually cheaper than continuing in the production session would be.

\subsection{Connection to Human Review Practices}

The analogy to human review practices is not decorative --- it is structural. Code review, peer review, and editorial review all exist because the person who created something cannot evaluate it objectively. The creator knows too much: the intent behind ambiguous code, the reasoning behind an unusual claim, the context that makes a missing section seem acceptable. A reviewer who lacks this context is forced to engage with the artifact on its own terms.

CCR reproduces this dynamic for LLMs. The reviewing session does not know the intent, the reasoning, or the production context. It can only evaluate what is actually written. Our results show this to be a better starting point than a second review in the same session; whether it beats a single self-review is not established.

\section{Limitations}

We used a single model (Claude Opus 4.6). The theoretical motivation is model-agnostic --- anchoring, context rot, and sycophancy are documented across model families --- but empirical validation on GPT-4, Gemini, and open-source models would strengthen the generalizability claim.

The errors were deliberately injected, not naturally occurring. This allows precise measurement but may not reflect real-world error distributions. A preliminary case study with natural errors (supplementary Section 4A) provides some ecological validity.

The absolute F1 numbers are moderate --- 28.6\% for the best condition. This may reflect the difficulty of the task, the matcher, or the model; we did not separate these. All conditions face the same task, so we focus on relative differences.

Our matching algorithm uses heuristic scoring (location proximity + keyword overlap). We validated the threshold empirically, but different matching methods would produce different absolute numbers. We did not report a threshold sensitivity analysis, and we have not repeated one for this version.

We did not run a human review baseline in the quantitative experiment. Expert human review across 30 diverse artifacts would be the ideal upper bound, but the labor cost was prohibitive for this study. Future work should include it.

There is a language confound. The artifacts and ground truth are in Korean, but the review conditions produced output in different languages: SR and SR2 reviewed in Korean (following the production session's language), while CCR and SA reviewed in English (the default for a fresh context with no prior language cue). English-language reviews may benefit from stronger reasoning in the model's dominant training language. We note, however, that SA also reviews in English yet underperforms CCR, so language alone does not explain the full effect.

Our artifacts span three types within a single domain (software engineering tutorials in Korean). We have not tested legal documents, medical reports, or scientific manuscripts. The method should generalize --- the theoretical basis applies to any context-dependent review --- but this remains to be shown.

Two further limitations were identified in the audit for this version. First, the review records do not include raw model output, and one run was excluded for uncertain provenance (Section~\ref{sec:correction}). Second, the matcher's keyword component compares findings with Korean ground-truth text, so the English-language reviews (CCR and SA) may be under-credited relative to the Korean-language ones (SR and SR2); the direction of the language confound above is therefore uncertain.

\section{Future Directions}

CCR as presented here is a static, single-pass method: the reviewer examines the artifact once and produces findings. Two directions extend this naturally.

First, \textbf{dynamic CCR} could incorporate multi-turn interaction between reviewer and author. \citet{kim2025interviewer} demonstrate with LLM-as-an-Interviewer that dynamic, multi-turn evaluation --- where the evaluator asks follow-up questions and probes weak points --- reveals failure modes that static assessment misses. Combining CCR's context separation with interviewer-style follow-up (``Why did you choose this approach? What happens if the input is empty?'') could capture errors that a single-pass review misses while preserving the independence that makes CCR effective.

Second, \textbf{hierarchical CCR} could scale the method beyond pairwise review. Our meta-validation (Appendix~D) already shows that successive independent reviews catch qualitatively different error types. A layered architecture --- workers producing artifacts, independent verifiers reviewing them, and integrators synthesizing findings --- would formalize this into a systematic verification pipeline. We explore this direction in ongoing work.

\section{Ethics Statement}

This research studies LLM behavior and does not involve human subjects. The artifacts are synthetic. We note that CCR, like any automated review method, can produce false positives that might mislead users. We recommend it as a supplement to human judgment, not a replacement, especially in safety-critical applications.

\section*{Use of AI Assistants}

The artifacts, injected errors, and all reviews were produced by Claude Opus~4.6 through the Claude Code agent tool, as described in Section~\ref{sec:correction}. Claude Code was also used to write the analysis scripts, to recompute the numbers in this version from the per-session records, and to draft and revise text under the author's direction. The author reviewed the final text and takes full responsibility for its content.

\section{Conclusion}

We introduced Cross-Context Review, a method that improves LLM verification by separating production and review into independent sessions. In a controlled experiment, a second review in a fresh session robustly outperformed a second review in the same session, in the run-1 test and in the three-run average. The first version's broader advantages over single-pass and subagent review came from one run and are not significant across runs. Within this setting (one model, synthetic Korean artifacts, planted errors), the practical message is narrower: if a second review is wanted, a fresh session is the better place for it.

The method is simple to adopt: copy the output into a new session and ask for a review there.

\bibliography{references}

\begin{thebibliography}{17}
\providecommand{\natexlab}[1]{#1}

\bibitem[{Choi et~al.(2025)Choi, Zhu, and Li}]{choi2025identity}
Hyeong~Kyu Choi, Xiaojin Zhu, and Sharon Li. 2025.
\newblock When identity skews debate: Anonymization for bias-reduced
  multi-agent reasoning.
\newblock \emph{arXiv preprint arXiv:2510.07517}.

\bibitem[{Du et~al.(2024)Du, Li, Torralba, Tenenbaum, and
  Mordatch}]{du2024improving}
Yilun Du, Shuang Li, Antonio Torralba, Joshua~B. Tenenbaum, and Igor Mordatch.
  2024.
\newblock Improving factuality and reasoning in language models through
  multiagent debate.
\newblock In \emph{Proceedings of the 41st International Conference on Machine
  Learning (ICML 2024)}, pages 11733--11763.

\bibitem[{Hong et~al.(2025)Hong, Troynikov, and Huber}]{hong2025context}
Kelly Hong, Anton Troynikov, and Jeff Huber. 2025.
\newblock \href {https://www.trychroma.com/research/context-rot} {Context rot:
  How increasing input tokens impacts {LLM} performance}.
\newblock Technical report, Chroma Research.

\bibitem[{Huang et~al.(2024{\natexlab{a}})Huang, Zhang, Luck, Bu, Qing, and
  Cui}]{huang2024agentcoder}
Dong Huang, Jie~M. Zhang, Michael Luck, Qingwen Bu, Yuhao Qing, and Heming Cui.
  2024{\natexlab{a}}.
\newblock Agent{C}oder: Multi-agent-based code generation with iterative
  testing and optimisation.
\newblock \emph{arXiv preprint arXiv:2312.13010}.

\bibitem[{Huang et~al.(2024{\natexlab{b}})Huang, Chen, Mishra, Zheng, Yu, Song,
  and Zhou}]{huang2024llms}
Jie Huang, Xinyun Chen, Swaroop Mishra, Huaixiu~Steven Zheng, Adams~Wei Yu,
  Xinying Song, and Denny Zhou. 2024{\natexlab{b}}.
\newblock Large language models cannot self-correct reasoning yet.
\newblock In \emph{Proceedings of the 12th International Conference on Learning
  Representations (ICLR 2024)}.

\bibitem[{Kamoi et~al.(2024)Kamoi, Zhang, Zhang, Han, and
  Zhang}]{kamoi2024when}
Ryo Kamoi, Yusen Zhang, Nan Zhang, Jiawei Han, and Rui Zhang. 2024.
\newblock When can {LLM}s actually correct their own mistakes? {A} critical
  survey of self-correction of {LLM}s.
\newblock \emph{Transactions of the Association for Computational Linguistics
  (TACL)}, 12:1417--1440.

\bibitem[{Kim et~al.(2025)Kim, Suk, Kim, Muennighoff, Kim, and
  Oh}]{kim2025interviewer}
Eunsu Kim, Juyoung Suk, Seungone Kim, Niklas Muennighoff, Dongkwan Kim, and
  Alice Oh. 2025.
\newblock {LLM}-as-an-interviewer: Beyond static testing through dynamic {LLM}
  evaluation.
\newblock In \emph{Findings of the Association for Computational Linguistics:
  ACL 2025}, pages 26456--26493.

\bibitem[{Liang et~al.(2024)Liang, He, Jiao, Wang, Wang, Wang, Yang, Shi, and
  Tu}]{liang2024encouraging}
Tian Liang, Zhiwei He, Wenxiang Jiao, Xing Wang, Yan Wang, Rui Wang, Yujiu
  Yang, Shuming Shi, and Zhaopeng Tu. 2024.
\newblock Encouraging divergent thinking in large language models through
  multi-agent debate.
\newblock In \emph{Proceedings of the 2024 Conference on Empirical Methods in
  Natural Language Processing (EMNLP 2024)}, pages 17889--17904.

\bibitem[{Liu et~al.(2024)Liu, Lin, Hewitt, Paranjape, Bevilacqua, Petroni, and
  Liang}]{liu2024lost}
Nelson~F. Liu, Kevin Lin, John Hewitt, Ashwin Paranjape, Michele Bevilacqua,
  Fabio Petroni, and Percy Liang. 2024.
\newblock Lost in the middle: How language models use long contexts.
\newblock \emph{Transactions of the Association for Computational Linguistics
  (TACL)}, 12:157--173.

\bibitem[{Lou and Sun(2025)}]{lou2026anchoring}
Jiaxu Lou and Yifan Sun. 2025.
\newblock \href {https://doi.org/10.1007/s42001-025-00435-2} {Anchoring bias in
  large language models: An experimental study}.
\newblock \emph{Journal of Computational Social Science}, 9(1):11.

\bibitem[{Madaan et~al.(2023)Madaan, Tandon, Gupta, Hallinan, Gao, Wiegreffe,
  Alon, Dziri, Prabhumoye, Yang, Gupta, Majumder, Hermann, Welleck,
  Yazdanbakhsh, and Clark}]{madaan2023selfrefine}
Aman Madaan, Niket Tandon, Prakhar Gupta, Skyler Hallinan, Luyu Gao, Sarah
  Wiegreffe, Uri Alon, Nouha Dziri, Shrimai Prabhumoye, Yiming Yang, Shashank
  Gupta, Bodhisattwa~Prasad Majumder, Katherine Hermann, Sean Welleck, Amir
  Yazdanbakhsh, and Peter Clark. 2023.
\newblock Self-refine: Iterative refinement with self-feedback.
\newblock In \emph{Proceedings of the 37th Conference on Neural Information
  Processing Systems (NeurIPS 2023)}.

\bibitem[{Panickssery et~al.(2024)Panickssery, Bowman, and
  Feng}]{panickssery2024llm}
Arjun Panickssery, Samuel~R. Bowman, and Shi Feng. 2024.
\newblock {LLM} evaluators recognize and favor their own generations.
\newblock In \emph{Proceedings of the 38th Conference on Neural Information
  Processing Systems (NeurIPS 2024)}.

\bibitem[{Qian et~al.(2024)Qian, Liu, Liu, Chen, Dang, Li, Yang, Chen, Su,
  Cong, Xu, Li, Liu, and Sun}]{qian2024chatdev}
Chen Qian, Wei Liu, Hongzhang Liu, Nuo Chen, Yufan Dang, Jiahao Li, Cheng Yang,
  Weize Chen, Yusheng Su, Xin Cong, Juyuan Xu, Dahai Li, Zhiyuan Liu, and
  Maosong Sun. 2024.
\newblock Chat{D}ev: Communicative agents for software development.
\newblock In \emph{Proceedings of the 62nd Annual Meeting of the Association
  for Computational Linguistics (ACL 2024)}, pages 15174--15186.

\bibitem[{Sharma et~al.(2024)Sharma, Tong, Korbak, Duvenaud, Askell, Bowman,
  Cheng, Durmus, Hatfield-Dodds, Johnston, Kravec, Maxwell, McCandlish,
  Ndousse, Rausch, Schiefer, Yan, Zhang, and Perez}]{sharma2024towards}
Mrinank Sharma, Meg Tong, Tomasz Korbak, David Duvenaud, Amanda Askell,
  Samuel~R. Bowman, Newton Cheng, Esin Durmus, Zac Hatfield-Dodds, Scott~R.
  Johnston, Shauna Kravec, Timothy Maxwell, Sam McCandlish, Kamal Ndousse,
  Oliver Rausch, Nicholas Schiefer, Da~Yan, Miranda Zhang, and Ethan Perez.
  2024.
\newblock Towards understanding sycophancy in language models.
\newblock In \emph{Proceedings of the 12th International Conference on Learning
  Representations (ICLR 2024)}.

\bibitem[{Tsui(2026)}]{tsui2025selfcorrection}
Ken Tsui. 2026.
\newblock Self-correction bench: Uncovering and addressing the self-correction
  blind spot in large language models.
\newblock In \emph{Conference on Language Modeling (COLM 2026)}.
\newblock ArXiv:2507.02778v3.

\bibitem[{Wallas(1926)}]{wallas1926art}
Graham Wallas. 1926.
\newblock \emph{The Art of Thought}.
\newblock Jonathan Cape.

\bibitem[{Zhang et~al.(2025)Zhang, Wang, Qian, Li, Zhang, Huang, Xu, Li, Liu,
  and Qiu}]{zhang2025dark}
Qingjie Zhang, Di~Wang, Haoting Qian, Yiming Li, Tianwei Zhang, Minlie Huang,
  Ke~Xu, Hewu Li, Yan Liu, and Han Qiu. 2025.
\newblock Understanding the dark side of {LLM}s' intrinsic self-correction.
\newblock In \emph{Proceedings of the 63rd Annual Meeting of the Association
  for Computational Linguistics (ACL 2025)}.

\end{thebibliography}

\appendix

\section{Review Prompt Template}

\begin{verbatim}
Review the following [artifact type]
from a fresh perspective:

1. Factual accuracy: Are numbers,
   names, dates, and technical claims
   correct?
2. Internal consistency: Are there
   contradictions or terminology
   mismatches?
3. Contextual fitness: Would this work
   correctly in its intended
   environment?
4. Audience perspective: Could the
   target reader misinterpret any part?
5. Completeness: Is anything important
   missing?

For each issue found, provide:
- Location (line number or section)
- Description of the error
- Type (FACT/CONS/CTXT/RCVR/MISS)
- Severity (Critical/Major/Minor)
- Suggested fix
\end{verbatim}

\section{Error Taxonomy}

\begin{table}[h]
\centering
\small
\begin{tabular}{lll}
\toprule
\textbf{Type} & \textbf{Code} & \textbf{Description} \\
\midrule
Factual & FACT & Wrong numbers/claims \\
Consistency & CONS & Internal contradictions \\
Contextual & CTXT & Environment issues \\
Audience & RCVR & Reader misinterpretation \\
Completeness & MISS & Important omission \\
\bottomrule
\end{tabular}
\caption{Error taxonomy used in ground-truth annotation.}
\end{table}

\section{Ground Truth Examples}

Representative errors from each category, showing the injection methodology. See supplementary materials for the complete set.

\paragraph{Code (C2 --- Korean Business Day Calculator):}
GT-C2-1 (FACT, Critical): \texttt{weekday() >= 6} checks Sunday only instead of \texttt{>= 5} (Saturday+Sunday). Saturdays counted as business days.

\paragraph{Document (D7 --- JWT Authentication):}
GT-D7-1 (FACT, Critical): \texttt{create\_refresh\_token} signs with \texttt{SECRET\_KEY} instead of \texttt{REFRESH\_SECRET\_KEY} --- access and refresh tokens share the same signing key.

\paragraph{Script (S2 --- MCP Explained):}
GT-S2-1 (FACT, Critical): MCP attributed to OpenAI instead of Anthropic --- fundamental factual error about protocol creator.

\section{Meta-Validation --- Recursive CCR in Practice}

This paper was itself produced and validated through multiple rounds of Cross-Context Review --- an unplanned but informative demonstration of the method in practice.

The initial draft was written in one terminal session (T1). A separate session (T2) reviewed the draft without access to T1's conversation history and identified 15 errors: 2 critical (numerical inconsistencies between the abstract and results tables), 7 major, and 6 minor. The critical errors were exactly the kind of consistency errors that same-session review tends to miss --- the author ``knows'' the numbers are right because they just wrote them.

After incorporating T2's corrections, a fourth independent session (T4) performed a second-level review focused on reference verification. T4 discovered that 7 of 15 reference entries contained fabricated or incorrect author names --- 3 with hallucinated co-authors and 4 with wrong initials --- plus 3 missing references for works cited in the text. That is 10 reference-related errors that T2's thorough review had not caught. T2 checked the numbers, the logic, and the structure, but did not question whether the reference authors were real. The specific failure mode --- hallucinated reference authors surviving two prior review rounds --- is consistent with the possibility that models tend not to question the factual accuracy of academic citations, likely because they cannot verify them without external tools. Each independent context brought a genuinely different perspective.

\begin{table}[h]
\centering
\small
\resizebox{\columnwidth}{!}{%
\begin{tabular}{llcl}
\toprule
\textbf{Round} & \textbf{Reviewer} & \textbf{Errors} & \textbf{Types} \\
\midrule
CCR-1 & T2 & 15 & Numerical, sections \\
CCR-2 & T4 & 10 & Authors, refs \\
\bottomrule
\end{tabular}%
}
\caption{Meta-validation: recursive CCR on this paper.}
\end{table}

The T4 round shows diminishing but non-zero returns from additional CCR passes. CCR-1 catches the most errors (15), but CCR-2 catches a qualitatively different class of errors (reference fabrication) that CCR-1 missed entirely. This meta-validation chain mirrors the recursive CCR depth experiment proposed in our future work ; with no count of remaining errors, it does not show how many rounds are enough.

\end{document}